\documentclass[twocolumn,conference]{IEEEtran}
\usepackage[T1]{fontenc}
\usepackage[latin9]{inputenc}
\usepackage{array}
\usepackage{float}
\usepackage{textcomp}
\usepackage{multirow}
\usepackage{amsmath}
\usepackage{graphicx}

\makeatletter

\providecommand{\tabularnewline}{\\}

\usepackage[caption=false,font=footnotesize]{subfig}

\makeatother

\begin{document}

\title{Detecting Zones and Threat on 3D Body for Security in Airports using Deep Machine Learning}

\author{\IEEEauthorblockN{Abel Ag Rb Guimaraes\\
}\IEEEauthorblockA{The G. Raymond Chang School of Continuing Education\\
Ryerson University\\
Toronto, ON M5B 2K3 Canada\\
Email: abel.guimaraes@ryerson.ca}\and \IEEEauthorblockN{Ghassem Tofighi\\
}\IEEEauthorblockA{Electrical and Computer Engineering Department\\
Ryerson University\\
Toronto, ON M5B 2K3 Canada\\
Email: gtofighi@ryerson.ca}}

\IEEEspecialpapernotice{}

\IEEEaftertitletext{}
\maketitle
\begin{abstract}
In this research, it was used a segmentation and classification method
to identify threat recognition in human scanner images of airport
security. The Department of Homeland Security's (DHS) in USA has a
higher false alarm, produced from theirs algorithms using today's
scanners at the airports. To repair this problem they started a new
competition at Kaggle site asking the science community to improve
their detection with new algorithms. The dataset used in this research
comes from DHS at https://www.kaggle.com/c/passenger-screening-algorithm-challenge/data.
According to DHS: \textquotedbl{}This dataset contains a large number
of body scans acquired by a new generation of millimeter wave scanner
called the High Definition-Advanced Imaging Technology (HD-AIT) system.
They are comprised of volunteers wearing different clothing types
(from light summer clothes to heavy winter clothes), different body
mass indices, different genders, different numbers of threats, and
different types of threats\textquotedbl{}. Using Python as a principal
language, the preprocessed of the dataset images extracted features
from 200 bodies using: intensity, intensity differences and local
neighbourhood to detect, to produce segmentation regions and label
those regions to be used as a truth in a training and test dataset.
The regions are subsequently give to a CNN deep learning classifier
to predict 17 classes (that represents the body zones): zone1, zone2,
... zone17 and zones with threat in a total of 34 zones. The analysis
showed the results of the classifier an accuracy of 98.2863\% and
a loss of 0.091319, as well as an average of 100\% for recall and
precision.\\
\emph{Keywords\textemdash Deep learning; DHS; Security;}\\
\end{abstract}

\IEEEpeerreviewmaketitle{}

\section{Introduction}

To be able to accurately identify objects camouflaged in the human
body through artificial intelligence is one of the big challenge facing
security today. There is, nowadays, a worldwide warning regarding
airport security. This happen because of the terrorism that threatens
the population or the great flow of foreigners that has been growing
year by year. New systems, programs and intelligence of protection
were created to help the teams that do security in the airports. According
to Bart Elias \cite{elias2012airport}, Transportation Security Administration
(TSA) has deployed 1,800 units Advanced Image Technology (AIT) throughout
the airports on USA and the acquisition cost per unit is about \$175,000.
To fully implement this system, the overall annual cost for purchasing,
installing, staffing, operating, supporting, upgrading and maintaining
sum to about 1.17 billion. Adding on it the Department of Homeland
Security\textquoteright s (DHS) in USA has a higher false alarm, produced
from their algorithms using today\textquoteright s scanners at the
airports. Although, the investment being made represents a significant
amount of money, it is not solving the security problem with regarding
to the detection of threats. The Security still needs to develop technology
to fix these algorithms to minimize the errors. Like others cleaners
of data, working with images need a special care in filter data that
should be a ground truth as possible. Peter Corke \cite{corke2011robotics},
discussed the difficulties in filter the noise in the images and in
segmenting the images in regions that contains the object. Errors
in the segmentation can mislead the security system. The ideal is
to reduce the noise and produce some threshold that correct the segmentation
regions. Considering this scenario, this paper work will present a
model to do the classification of the regions of the body\textasciiacute s
images using supervised machine learning to identify the threats hired
in the bodies. The results will be analyzed to verify the true or
false of the information collected in the model. The objectives proposed
in this research are: Produce an algorithm that fractionated the human
body image in regions to be able to identify the body\textquoteright s
region correctly. Use a supervised classification to recognize the
human body\textquoteright s regions. Use the training and test dataset
to identify the true or false of the detection, for measuring their
accuracy, recall and precision. Propose improvements in the algorithm
for implementation.

\section{Previous Work and Literature}

Bart Elias \cite{elias2012airport} described how Transport Security
Administration (TSA) implement the new AIT units, using new millimeter
wave images, that\textquoteright s substitute the X-ray backscatter
units, exposing the difference between those units and the risk of
both systems in terms of health concern. It explains and describe
Automated Target Recognition (ATR), also the scanners effectiveness
measure regarding accuracy and the trade-off between detection and
false alarm. David Powers \cite{powers2011evaluation} evaluated measures
including Recall, Precision, F-Measure, Rand Accuracy and how the
biases transform the concepts of the results, demonstrating the elegant
connection of Informedness, Markedness, Correlation and Significance,
as well providing their intuitive relationships with Recall and Precision.
Sauvola \cite{sauvola2000adaptive} demonstrated a method of binary
image with local and multiple threshold evaluation, instead of a global
threshold. Those measures used formulas that take the window that
surround of every pixel to take the threshold based on mean and standard
deviation of the local neighborhood. Using a Hybrid switch technique,
one method is based on histogram and other based on Soft decision
(SDM), to produce a threshold using interpolation for a better result.
Robinson \cite{robinson2004efficient} used a morphological reconstruction
by dilation, that is different from the basic morphological dilation,
where high-intensity values are replaced by nearby low-intensity values.
In contrast, it uses two images, one image called \textquotedblleft seed\textquotedblright ,
that specifies how far values can be spread, and a \textquotedblleft mask\textquotedblright{}
image, which allow each pixel to have a maximum value limiting the
spread of high-intensity values. In his book, Peter \cite{corke2011robotics}
discussed about many techniques of image process Like monodic function
(normalization, threshold, gamma), diadic functions (temporal smoothing),
spatial (morphological), features (histograms), shape change (rotation,
pyramid scale, ROI) among others. The language and the explanation
that were used in this book facilitated the understanding. In this
article, Sarraf and Tofighi \cite{sarraf2016deep} exposed a CNN deep
learning data model using fMRI images to identify regions that has
a pattern of Alzheimer\textquoteright s Disease. It explained how
the convolution and pull data work in the process of images, in which
is obtained a classification. In addition, they showed the result
using Receiver Operating Characteristics (ROC) curve, taking about
96.86\% of the accuracy. Alex Krizhevsky \cite{krizhevsky2012imagenet}
used a neural network with 60 million parameters and 650,000 neurons,
their method consists of five convolution layers followed by max-pooling
using a final 1000-way softmax. To make training faster, they used
non-saturating neurons and a very efficient GPU implementation of
the convolution operation. To reduce overfitting in the fully-connected
layers was used a regularization method called \textquotedblleft dropout\textquotedblright .
LeCun \cite{lecun1998gradient}\cite{lecun1998efficient} introduced
the gradient-based learning for document recognition. It explained
most of all techniques in Neural Network principally CNN, that imposed
convolution learning in the layer steps, in that way it has invariance
of shift, scale and distortion. Leon Bottou \cite{bottou2012stochastic},
strongly advocates to use the stochastic back-propagation method to
train neural networks. It explains why SGD is a good learning algorithm
to train large datasets. Those recommendations collaborated with this
research. Bengio \cite{bengio2012practical} prepared a practical
guide recommendation to adjust common many hyper-parameters, based
on back-propagated gradient, as well did questions about training
difficulties observed in deep machine learning. Kumar \cite{chellapilla2006high}
showed the basic ideas of how CNN was improved when it implements
the interface with GPU. TensorFlow \cite{tensorflow2015-whitepaper}
is a model that represents and uses nodes as a data-flow graph, those
nodes operates individual mathematical such as matrix multiplication,
convolution, min and max and many more. With this approach user can
compose layers using a high-level scripting or in a graph interface.
Caffe \cite{jia2014caffe} like TensorFlow, it is an open-source framework
to access deep architectures. It uses CUDA for GPU computation that
is implemented with C++, it can use to do an interface with Python/Numpy
and Matlab. It is possible to process over 40 million images a day
on a single K40 or Titan GPU (\ensuremath{\approx} 2.5 ms per image).
It also separates model network from actual implementation. Bishop
\cite{bishop2006pattern} presented in his book most of all techniques
to learn pattern recognition, it goes to the basic probability showing
all principal model like regression, svn and others. It's a good book
to understand about pattern recognition and machine learning in all
aspects. In his article, Davis \cite{davis2006relationship} explained
about both ROC curves and Recall Precision Graphs model, moreover
he showed the difference between classes comparison. When its comparing
two classes in the graph, class one showed a better result in ROC
curve, than the class two; when change the graph for PR, class two
has better result. In the Python world, Wall \cite{walt2011numpy}
developed the NumPy arrays. NumPy arrays are the standard representation
for numerical data according to the community. Numpy arrays can implement
efficient numerical computations for a high-level compute language
using Python. There are three techniques that improve performance:
vectorizing calculations, avoiding copying data in memory, and minimizing
operation counts.

\section{Methodology}

\subsection{Dataset}

The dataset used in this research comes from DHS and Kaggle competition
at: https://www.kaggle.com/c/passenger-screening-algorithm- challenge/data/
. According to DHS: \textquotedblleft This dataset contains a large
number of body scans acquired by a new generation of millimeter wave
scanner called the High Definition-Advanced Imaging Technology (HD-AIT)
system. They are comprised of volunteers wearing different clothing
types (from light summer clothes to heavy winter clothes), different
body mass indices, different genders, different numbers of threats,
and different types of threats\textquotedblright . It will start with
a small sample of the body\textquoteright s scanned separating the
data into two parts, to be used as a training set and testing set.
After that, we can use more samples of the huge database to get better
accuracy. For the stage1 of Kaggle Competition we have a total of
929 bodies with threat to be analyzed. Here, in this research, we
will limit for a sample of 200 bodies to be analyzed; It will produce
a total of 287660 sliced images of the bodies.

\subsection{Approach}

Here is the block diagram of this paper. It tries to summarize all
the steps needed to describe the algorithms in every step, those algorithms
are writing using python language. A first approach is to divide the
bodies in 17 regions that can contain threat or cannot contains threats.

\begin{figure}[htbp]
\begin{centering}
\textsf{\includegraphics{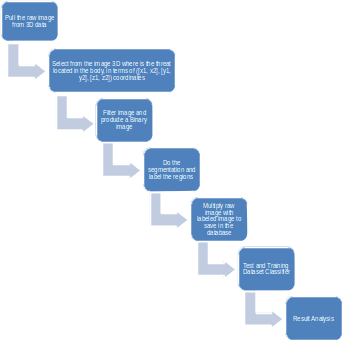}}
\par\end{centering}
\caption{Approach block diagram}
\end{figure}

\subsubsection{Pull the raw image from 3D data}

This step loads the data files, the 200 bodies from TSA database,
in a extension of a3d file. The scanned body\textquoteright s will
be loaded one by one to produce labels regions. Each scanned sliced
view has the length of the body through a z axis coordinate. Besides,
TSA produces more 3 types of file scanned body: one that is scanned
rotated around the z axis and saved with the extension aps; second
in a mixed of ad3 and aps and; the last one is a high resolution of
the body. In this research was used both the aps and a3d. The file
a3d can be worked with the sliced plane (x, y) image of the body and
the aps file, it was used to locate the threat's body.

\subsubsection{Select from the image 3D where is the threat located in the body,
in terms of ({[}x1, x2{]}, {[}y1, y2{]}, {[}z1, z2{]}) coordinates}

This step pulls the threats from the dataset TSA that uses the file
with extension aps. Every body will be rotated to localize the threat
to pull the coordinates of the threat. Those coordinates will be saved
in a table with the extension of csv file (table 1). An algorithm
from Kaggle was modified to plot the bodies in a grid to determine
the threat coordinates (x, y, z) (as shown at the Figure 2);

\begin{figure}[htbp]
\begin{centering}
\textsf{\includegraphics[scale=0.3]{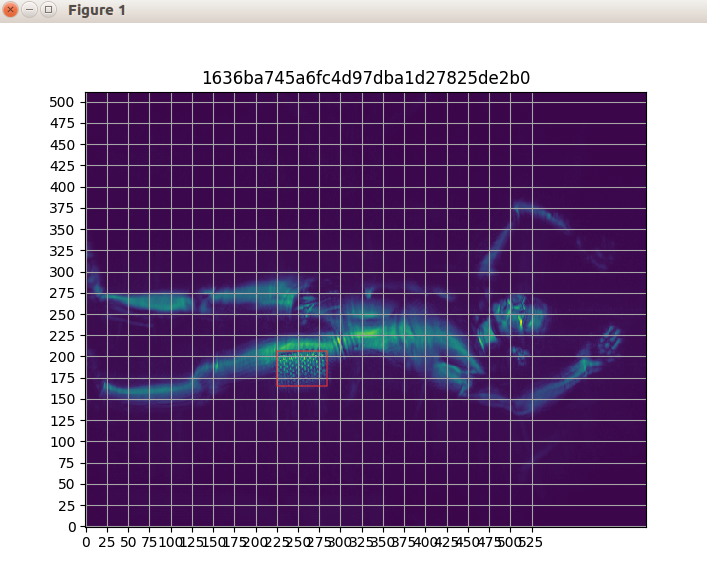}}
\par\end{centering}
\caption{Algorithm show a body with a threat located at red}
\end{figure}

\begin{table}[htbp]
{\tiny{}\caption{Label zones and threats}
}{\tiny \par}
\centering{}{\tiny{}}%
\begin{tabular}{|c|c|c|c|c|c|c|>{\centering}m{1cm}|}
\hline 
{\tiny{}body\_Id} & {\tiny{}z\_start} & {\tiny{}z\_stop} & {\tiny{}zone} & {\tiny{}x\_start} & {\tiny{}x\_stop} & {\tiny{}y\_start} & {\tiny{}y\_stop}\tabularnewline
\hline 
\hline 
{\tiny{}00360} & {\tiny{}88} & {\tiny{}127} & {\tiny{}14} & {\tiny{}290} & {\tiny{}340} & {\tiny{}350} & {\tiny{}373}\tabularnewline
\hline 
\hline 
{\tiny{}0043d} & {\tiny{}444} & {\tiny{}475} & {\tiny{}1} & {\tiny{}354} & {\tiny{}435} & {\tiny{}260} & {\tiny{}288}\tabularnewline
\hline 
\hline 
{\tiny{}0043d} & {\tiny{}75} & {\tiny{}139} & {\tiny{}14} & {\tiny{}125} & {\tiny{}195} & {\tiny{}200} & {\tiny{}250}\tabularnewline
\hline 
\hline 
{\tiny{}0043d} & {\tiny{}208} & {\tiny{}279} & {\tiny{}9} & {\tiny{}130} & {\tiny{}150} & {\tiny{}275} & {\tiny{}300}\tabularnewline
\hline 
\hline 
{\tiny{}00504} & {\tiny{}158} & {\tiny{}214} & {\tiny{}8} & {\tiny{}320} & {\tiny{}375} & {\tiny{}189} & {\tiny{}236}\tabularnewline
\hline 
\end{tabular}{\tiny \par}
\end{table}

\subsubsection{Filter image and produce a Binary image}

This step filters every slice of the body to get the Region of Interest
(ROI). The scanned image comes with a lot of noise. To filter that
noise and obtain the ROI result, the image is submitted to a threshold
filter, this threshold cleans the image to produce the segmentation.
Using a Gaussian convolution, the segmented image will be normalized,
this means, this image will be cleaned from the noise and will produce
a better result to obtain the ROI. After that, the segmented image
will be imposed to a dilatation process that join the segmented parts
of the same object, cleaning the background. Images below on table
2 exposes the flow of raw image before and after the threshold filter:

\begin{table}[htbp]
\caption{Process of cleaning and producing a binary image}
\centering{}%
\begin{tabular}{|c|c|c|}
\hline 
Raw Image & Clean process & Binary image\tabularnewline
\hline 
\hline 
\includegraphics[scale=0.3]{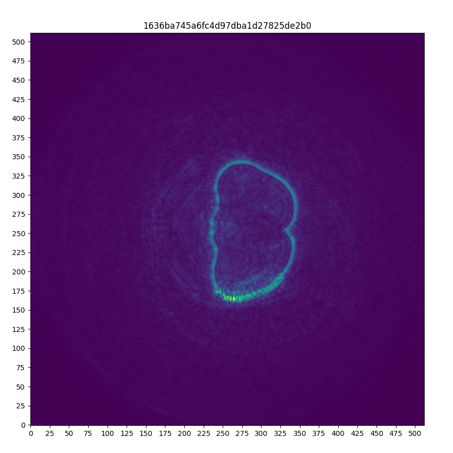} & \multicolumn{1}{c|}{$\mathbf{c}[u,v]=\begin{cases}
0 & \mathbf{I}[u,v]<T\\
1 & \mathbf{I}[u,v]\geq T
\end{cases}\forall(u,v)\in\mathbf{I}$} & \includegraphics[scale=0.3]{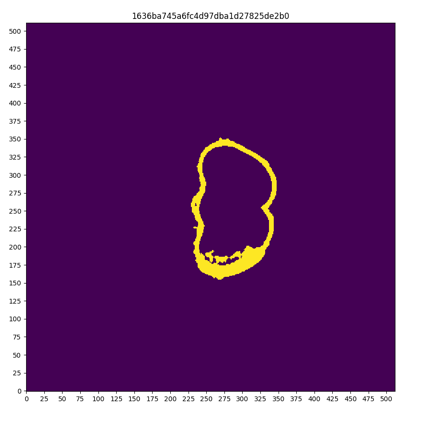}\tabularnewline
\hline 
\end{tabular}
\end{table}

\subsubsection{Do the segmentation and label the Regions}

\begin{figure}[htbp]
\begin{centering}
\includegraphics[scale=0.5]{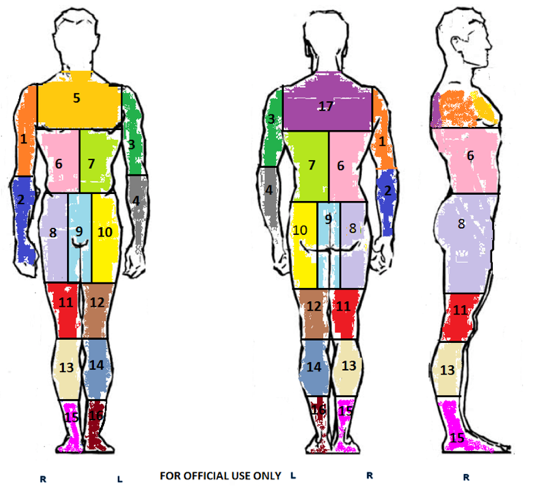}
\par\end{centering}
\caption{17 zones of the bodies from kaggle TSA competition}
\end{figure}
On this step, the segmented image will be identified (labeled) in
the right zone of the body, tracking the ROI through z axis. It
starts taking a look at every ROI in the sliced image, separating
them and combining in the right position on the z axis, it will produce
a cloud of points of each zone of the human body. In that way, was
produced 8 zones (through z axis) that represents the total human
body. On the other hand, considering the x axis, was produced 16 zones,
right and left side in a symmetric way and including a specific zone
(zone 9 in figure 3) were completed the 17 zones.

\begin{figure}[htbp]
\begin{centering}
\includegraphics[scale=0.5]{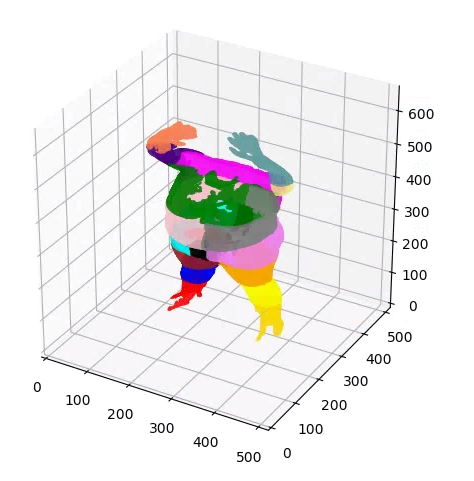}
\par\end{centering}
\caption{Final result of segmentation the body now has 17 }
\end{figure}
To produce the right and left regions, was used a mask to guide the
right ROI to the left ROI. Adding on this idea, the body was segmented
on the z axis in a proportion of the total length of the body. In
this way, the mask should guide in (x, y) position as well as the
z axis position. The figure 3 shows the 17 zones of the human body.
Each zone can have threat or not, so in this research, to classify
the threat, were considered 34 zones, that is double the 17 zones
of the body, it will be explain better on next steps.

The figure 4 shows the cloud of points of the 17 zones, that was segmented
using the proportion of the height of each body.
\subsubsection{Multiply raw image with labeled image to save in the database}

This step prepares the final image and save in a template directory
to be loaded to a Digits interface. A Multiplication process get the
labeled image and multiply with the raw image to get the real intensity
of the image, it focuses on the ROI of the image. The table 3 shows
the flow of the process: 

\begin{table}[tbph]
\caption{Process of Multiply image}
\centering{}%
\begin{tabular}{|c|c|}
\hline 
\multirow{1}{*}{Raw Image} & Binary image\tabularnewline
\hline 
\hline 
\includegraphics[scale=0.6]{figs/slice_image_raw} & \includegraphics[scale=0.6]{figs/slice_image_binary}\tabularnewline
\hline 
\multicolumn{2}{|c|}{Multiply process}\tabularnewline
\hline 
\multicolumn{2}{|c|}{$\boldsymbol{O}[u,v]=f\left(\boldsymbol{I}_{1}[u,v],\boldsymbol{I}_{2}[u,v]\right),\forall(u,v)\in\boldsymbol{I}_{1}$}\tabularnewline
\hline 
\multicolumn{2}{|c|}{Final image}\tabularnewline
\hline 
\multicolumn{2}{|c|}{\includegraphics[scale=0.6]{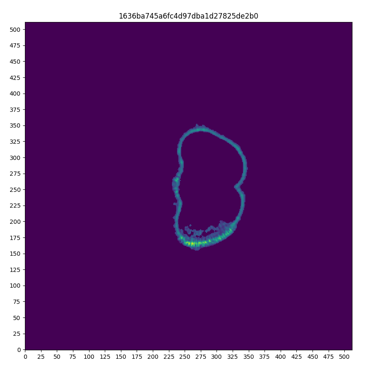}}\tabularnewline
\hline 
\end{tabular}
\end{table}
After that, all sliced image are saved in directories zones and threat
zones as showed at figure 5:

\begin{figure}[htbp]
\begin{centering}
\includegraphics[scale=0.2]{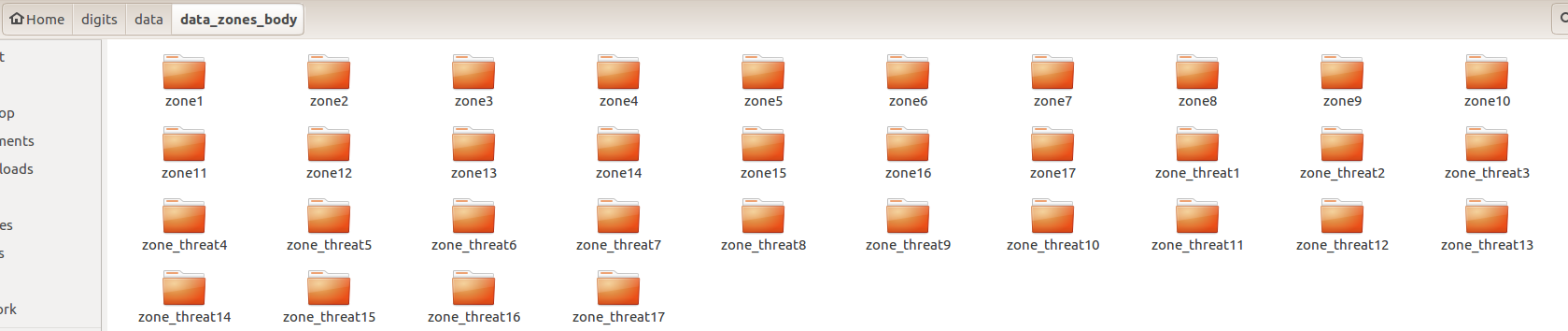}
\par\end{centering}
\caption{Directory division that contain zones}
\end{figure}

\subsubsection{Test and Training Dataset Classifier}

After have all the zones labeled and written in a file, it can be
loaded to fit a classification model. The training and test data will
be separated in two portion: 60\% and 40\%. The training data will
be with 60\% of the total data and the test will take the rest of
portion. The trained data was submitted in a configured model
called CNN, with the solver type Stochastic Gradient Decent (SGD)
using the network AlexNet and frame work Tensorflow. The AlexNet network
consists of 7 layers, 5 convolutions by 5 max-pools and 2 regression
layers at the end. The figure 6 shows the flow of AlexNet work:

\begin{figure}[htbp]
\begin{centering}
\includegraphics[scale=0.2]{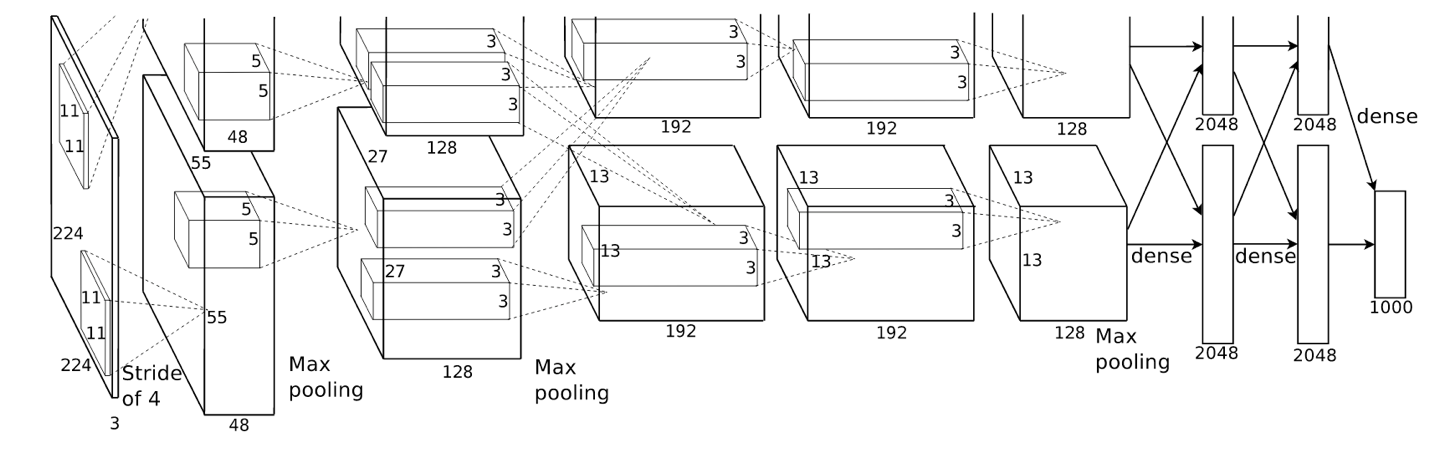}
\par\end{centering}
\caption{AlexNet \cite{krizhevsky2012imagenet} flow of convolution and pulling
steps from the original paper.}
\end{figure}

AlexNet \cite{krizhevsky2012imagenet} has some advantages in the
architecture that can be pointed:
\begin{enumerate}
\item It uses ReLU Non-linearity for non-saturating non-linearity, in this
way, the training time of the gradient descent is much faster than
the saturated non-linearitys.
\item It reduces the over-fitting using data augmentation and preserving
transformation using label. Those transformation of the data do not
need to be saved on a disk.
\item It dropout predictions, that is a very successful way to reduce test
errors.
\end{enumerate}
About the solver Stochastic Gradient Decent (SGD), Leon Bottou \cite{bottou2012stochastic}
point that the Big data size of learning is a bottleneck, so SGD can
perform very well with train time.

Besides AlexNet, there are more libraries that also can be use to
do machine learning:
\begin{itemize}
\item Digits from Nvidia, was used in this research, it is a IDE with a
frame-network like caffe, torch, tensorflow and can choose models
from their library. In this way, it can be chosen a particular model
to training the dataset.
\item Other libraries like scikit-learn that generate graphics like ROC,
it can be access with this link http://scikit-learn.org/.
\end{itemize}

\subsubsection{Result Analysis}

After the AlexNet process and CNN learning, a function model was produced
here. This model could classify images in zones.
Once the model was trained, we checked the accuracy on unseen test
data. This was done using the confusion matrix evaluating the recall
and precision of the trained model, comparing the predict class with
the actual class. After obtained the metric it could use the ROC curve
to compare the classifier. The results were saved and the curves was
plotted using sckit learn python library.

\section{Results}

The database created here used a LMDB format and was loaded to produce
a total of 287660 images that was submitted to training, validation
and test. Below at figure 7, its shows the result of the split images
(from Digits). Analyzing the 3 histograms, we can see that the data
is unbalanced, especially with the threats. The Digits used a total
of 34 zones (with threat or not). The training zones were fitted with
a total of 172596 images (60\%), using splitted images to train; to
validation were used splitted images in a total of 57532 (20\%) and
for test was used 57534 images (20 \%) and a format of 256x256 PNG
image.

\begin{figure}[htbp]
\begin{centering}
\textsf{\includegraphics[scale=0.5]{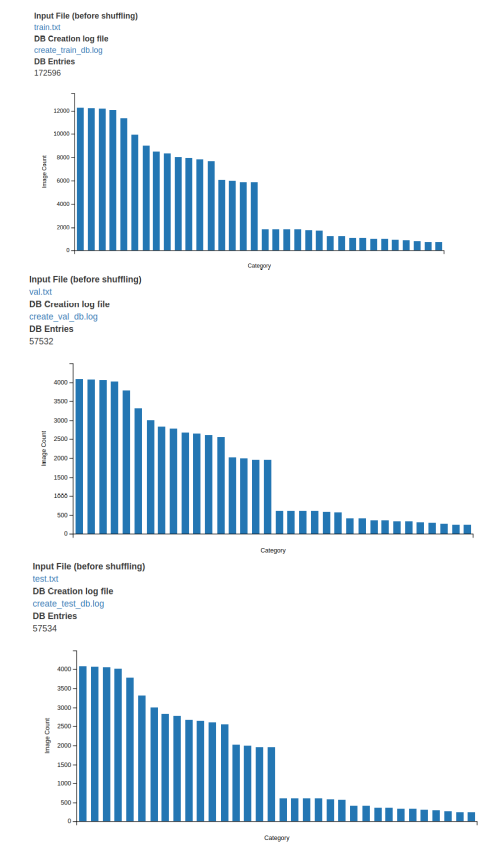}}
\par\end{centering}
\caption{Histograms of images zones from DB train, Db Validation and DB Test.}
\end{figure}
The training, Validation and test were performed in a Ubuntu machine
that has 16GB of memory, a I7 processor and a Nvidia GPU of 6GB. After
30 epochs, the model shows at the figure 8, an average accuracy for
the validation data of 98.2863\% and a loss of 0.091319. This was
obtained with a shuffled data using AlexNet with Tensor-flow. The
data transformation used a subtract mean image and data augmentation
to do the flipping for the threat zones and produce more balance data,
as well, using a contrast of 0.8 factor in the images.

\begin{figure}[htbp]
\begin{centering}
\textsf{\includegraphics[scale=0.28]{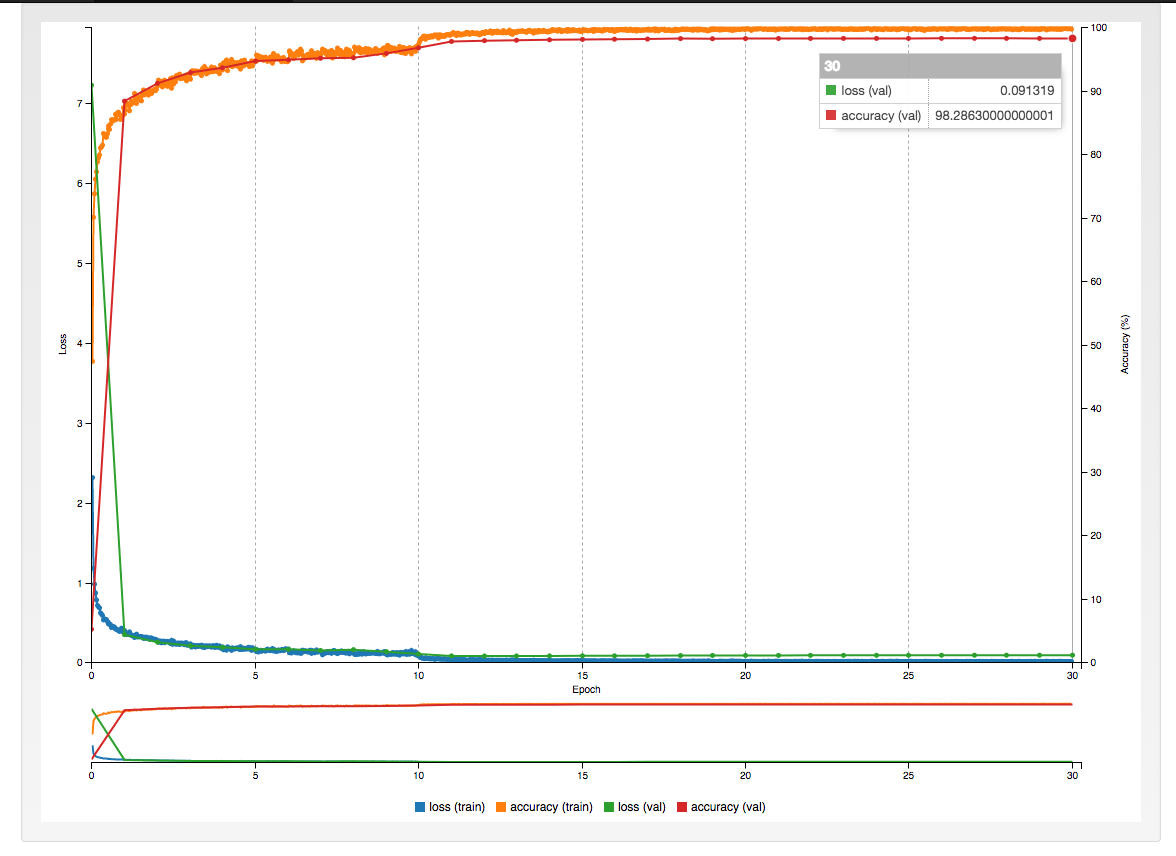}}
\par\end{centering}
\caption{Graph of Loss and accuracy with 30 epochs.}
\end{figure}
After process the train with Digits, to test the model, we can use
only 1 image. In figure 8, was used the Classify One Image to show
Visualizations and Statistics when its selected. When running this
process it will generate plots of weights from the responses of the
network flow input image. Follow below the figure 9, it shows example
output from the first layer.

\begin{figure}[htbp]
\begin{centering}
\includegraphics[scale=0.2]{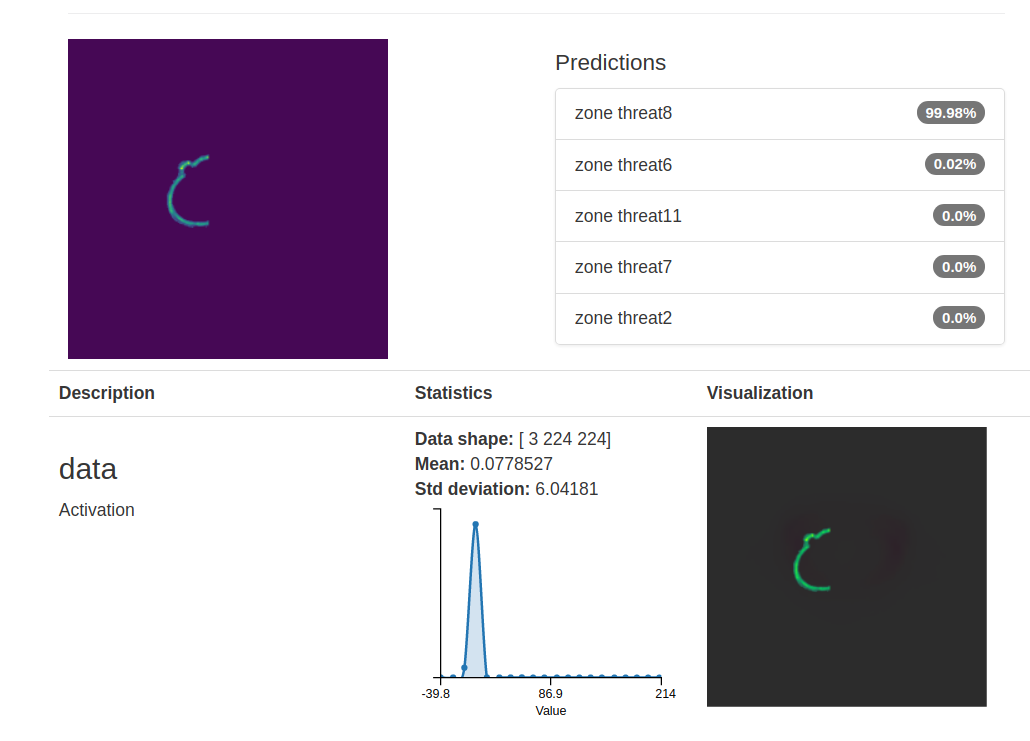}
\par\end{centering}
\caption{Classifying on image with statistics.}
\end{figure}

DIGITS plots statistical information in every convolution layer parameter,
this include frequency, mean and standard deviation. This is a helpful
tool to understand the results of convolution response applied in
the input image. The classification results displayed at the figure
10 shows a response from the first convolution layer including the
weights, activations and statistical information. \textquotedblleft In
a well-trained network, smooth filters without noisy patterns are
usually discovered. A smooth pattern without noise is an indicator
that the training process is sufficiently long, and likely no over-fitting
occurred. In addition, visualization of the activation of the network\textquoteright s
features is a helpful technique to explore training progress. In deeper
layers, the features become sparser and localized, and visualization
helps to explore any potential dead filters (all zero features for
many inputs).\textquotedblright{} (Sarraf and Tofigh \cite{sarraf2016deep})
The smoothness of the convolution images shows below follow the same
pattern.

\begin{figure}[htbp]
\begin{centering}
\includegraphics[scale=0.2]{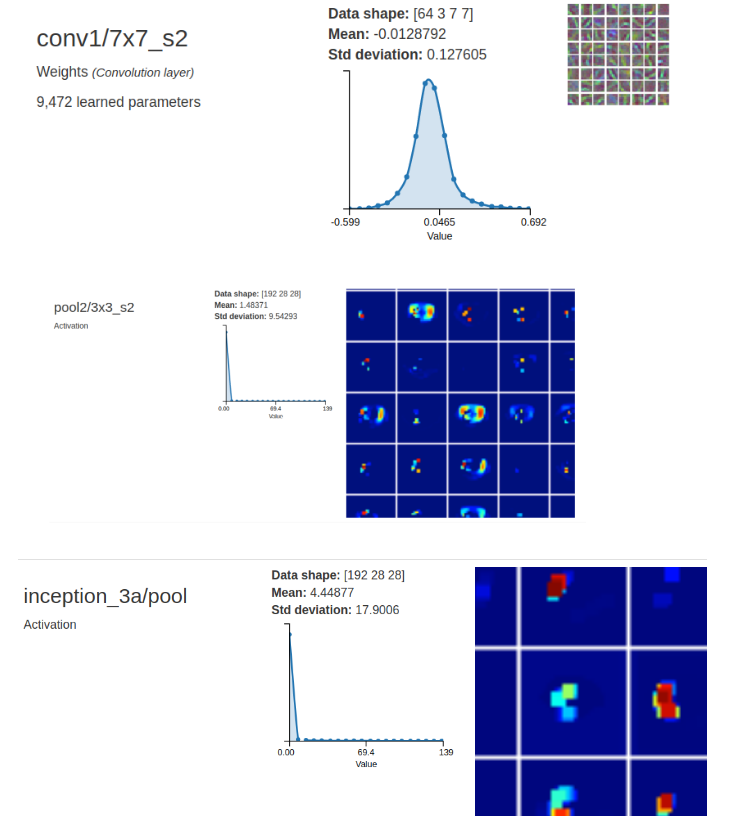}
\par\end{centering}
\caption{Statistics of the convolution, activations and weights.}
\end{figure}

To get a real result, it was used a different test data to enforce
the validation of the model. The figure 11 shows the tops accuracy
zones, it obtained a Top-1 accuracy of 99.31\% and Top-5 accuracy
of 100\%. 

\begin{figure}[htbp]
\begin{centering}
\includegraphics[scale=0.25]{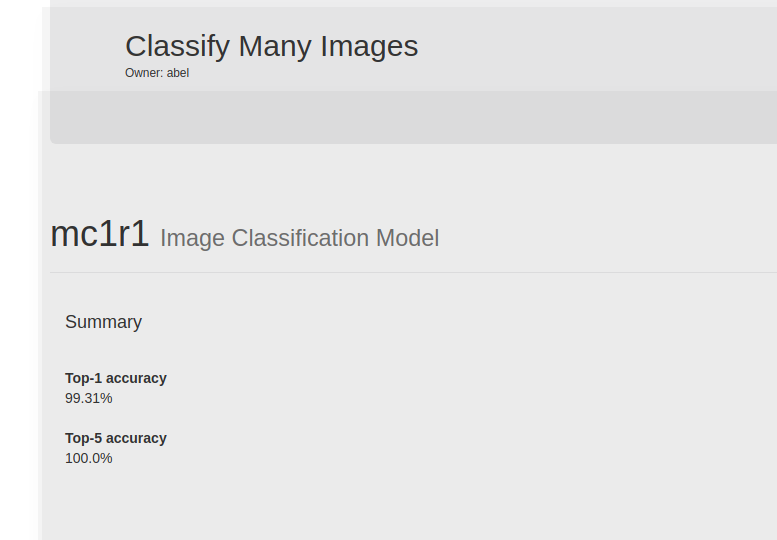}
\par\end{centering}
\caption{Classify many images summary.}
\end{figure}

After that, the confusion matrix at figure 12 shows all the accuracies
by class zones, most of them has results upper to 98\%.

\begin{figure}[H]
\begin{centering}
\includegraphics[scale=0.3]{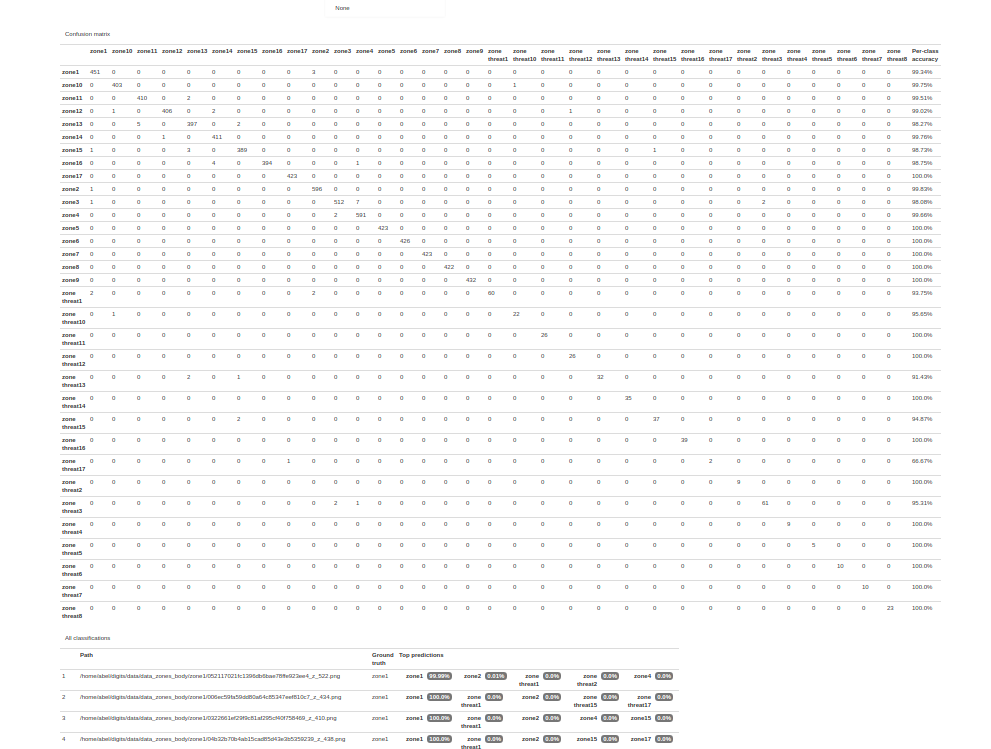}
\par\end{centering}
\caption{Confusion Matrix using classify many images summary from Digits.}
\end{figure}

Analyzing the 2 figures bellow: figure 13 and figure 14, we can obtain
Recall and Precision of the model. Recall represents the true positives
recovered from the true positives of the total sample, and Precision
represents the true positives of the answer produced from the model.
Starting with the analyses of Recall and Precision based on the results
of Classifying Many Images Test and their Confusion Matrix, we can
see the 2 graphs below. The figure 13 shows an average of 100\% in
Recall of the images as well as 100\% in precision.

\begin{figure}[H]
\begin{centering}
\includegraphics[scale=0.5]{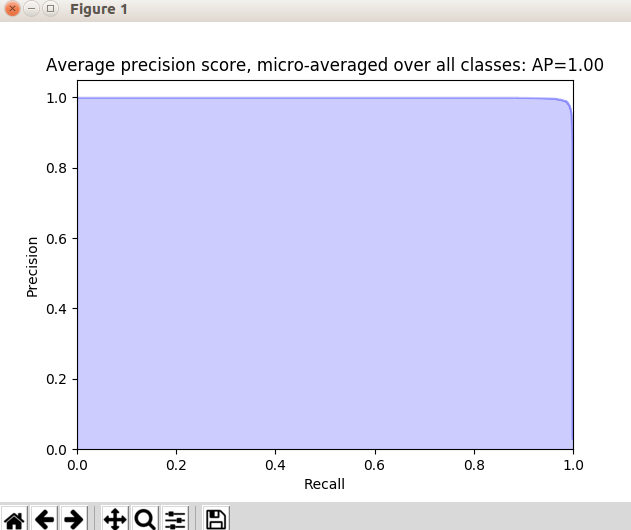}
\par\end{centering}
\caption{Average of Precision and Recall of bodies zone and zones with threats.}
\end{figure}
Figure 14 shows in more details all Precision and Recall by zones.
It can be analyzed more precisely about how much true positives are
recovering from the test images, as well as, the precision of this
true positives. It can analyze all zones individually and most of
them has a 100\% in area for both rates.

\begin{figure}[H]
\begin{centering}
\includegraphics[scale=0.25]{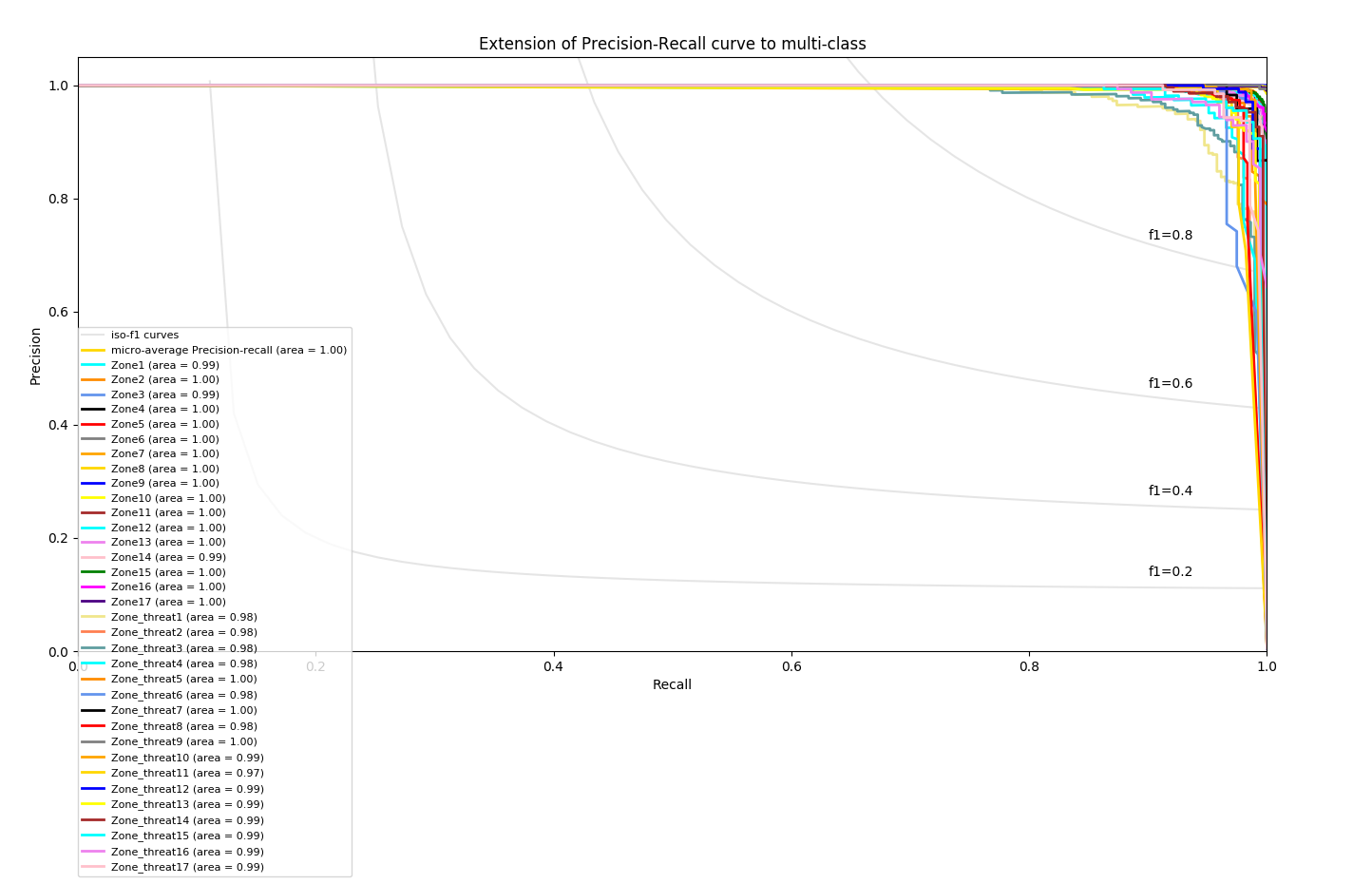}
\par\end{centering}
\caption{Precision and Recall of all zones and zones with threats.}
\end{figure}

\section{Conclusions}

Considering the proposal in the Introduction of this research, this
paper work intended to present a model to do the classification of
the regions of the body's images using supervised machine learning
to identify the threats hired in the bodies. Moreover, produce an
algorithm that fractionated the human body image in regions to be
able to identify the body\textquoteright s region correctly. In this
paper work, it was developed a method to do the segmentation of the
body in zones. It was presented 34 zones (considering 17 zones with
left and right regions plus the threats zones) to do the learning
process. Using robust module CNN with Stochastic Gradient Decent (SGD)
and AlexNet, the model could learn the zones with threat or not. Combining
with filters, the model could classify those zones of the bodies with
a robust result that can help TSA agency security in all airports
of North America and world. A big data files was used to train this
deep learning to extract features using convolution neural network
that can produce a faster learning provide by Nvidia Digits framework.
One important point to be considered here is the fact that the model
worked for the trained images of the sample used, it was not considered
any external image.

Another point is that the result produced was highly accurate for
predict zones, it can indicate an overfitting of learning. To prove
that, it is necessary to do more test using external bodies, besides
that the model need a balance data, that is, more information about
the threats. This research showed in average an accuracy of 100\%
of the bodies zones, as well as, a recall and precision with the same
100\% result for the most of zones. For future work, it is proposed
here the follow: one classifier that combine two different models:
one model that classify only the zones of the body and another model
that only classifies the threat or not, increase the body images from
external samples for produce a balance data.

\section{Acknowledgements}

This project is done as a part of Chang School Big Data Program.

I would like to express my gratitude towards Dr. Ghassem Tofigh, Ph.D.
in Electrical and Computer Engineering, (Computer Vision and Pattern
Recognition) from Ryerson University, and all fellows and professors
from The G. Raymond Chang School of Continuing Education and Big Data
Course.
The algorithm for this research is available at https://github.com/abelguima/ryerson-capstone-CKME136

\appendices{}

\bibliographystyle{../../IEEEtran}
\bibliography{../../IEEEabrv,../../IEEEexample_work}

\end{document}